\def\year{2020}\relax
\documentclass[letterpaper]{article}
\usepackage{aaai20}
\usepackage{times}
\usepackage{helvet}
\usepackage{courier}
\usepackage{url}
\usepackage{graphicx}
\usepackage{algorithm}
\usepackage{algorithmic}
\usepackage{amsmath}
\usepackage{subfigure}

\usepackage{soul}

\newcommand{\as}[1]{\textit{AutoShrink}}
\DeclareMathOperator*{\argmax}{arg\,max}

\title{AutoShrink: A Topology-aware NAS for Discovering Efficient Neural Architecture}
\author{Tunhou Zhang,\textsuperscript{\rm 1} 
Hsin-Pai Cheng,\textsuperscript{\rm 1} 
Zhenwen Li,\textsuperscript{\rm 2}
Feng Yan,\textsuperscript{\rm 3}
Chengyu Huang,\textsuperscript{\rm 4}
Hai Li,\textsuperscript{\rm 1}
Yiran Chen\textsuperscript{\rm 1}\\
\textsuperscript{\rm 1}ECE Department, Duke University, Durham, NC 27708\\
\textsuperscript{\rm 2}Institute of Computational Linguistics,
Peking University, Beijing, China \\
\textsuperscript{\rm 3}CSE Department, University of Nevada, Reno, NV 89557\\
\textsuperscript{\rm 4}Department of Electronic Engineering, Tsinghua University, Beijing \\
\{tunhou.zhang,dave.cheng,hai.li,yiran.chen\}@duke.edu,\\
lizhenwen@pku.edu.cn,
fyan@unr.edu,
huangcy16@mails.tsinghua.edu.cn
}
\begin{document}
\maketitle
\begin{abstract}
Resource is an important constraint when deploying Deep Neural Networks (DNNs) on mobile and edge devices. 
Existing works commonly adopt the cell-based search approach, which limits the flexibility of network patterns in learned cell structures.
Moreover, due to the topology-agnostic nature of existing works, including both cell-based and node-based approaches, the search process is time consuming and the performance of found architecture may be sub-optimal.
To address these problems, we propose \as{}, a topology-aware Neural Architecture Search (NAS) for searching efficient building blocks of neural architectures.
Our method is node-based and thus can learn flexible network patterns in cell structures within a topological search space.
Directed Acyclic Graphs (DAGs) are used to abstract DNN architectures and progressively optimize the cell structure through edge shrinking.
As the search space intrinsically reduces as the edges are progressively shrunk, \as{} explores more flexible search space with even less search time. 
We evaluate \as{} on image classification and language tasks by crafting \textit{ShrinkCNN} and \textit{ShrinkRNN} models.
ShrinkCNN is able to achieve up to 48\% parameter reduction and save 34\% Multiply-Accumulates (MACs) on ImageNet-1K with comparable accuracy of state-of-the-art (SOTA) models.
Specifically, both ShrinkCNN and ShrinkRNN are crafted within 1.5 GPU hours, which is $7.2\times$ and $6.7\times$ faster than the crafting time of SOTA CNN and RNN models, respectively.
\end{abstract}
\section{Introduction}
Neural Architecture Search (NAS) emerged only in recent years but has already demonstrated great strength in designing neural architectures automatically. 
Many research works show that neural architectures obtained from NAS surpass the performance of the hand-crafted counterpart for challenging tasks, such as computer vision and natural language processing~\cite{liu2018darts,long2015learning,tan2019mnasnet,tan2019efficientnet}.
The study on NAS methods in early stage concentrated on seeking large-scale neural architectures that can provide record breaking performance~\cite{long2015learning}. 
As computational power and computing time are not taken into consideration, the redundancy is inevitable for the models obtained by these methods.

\begin{figure}[t]
\centering
\includegraphics[width=0.95\columnwidth]{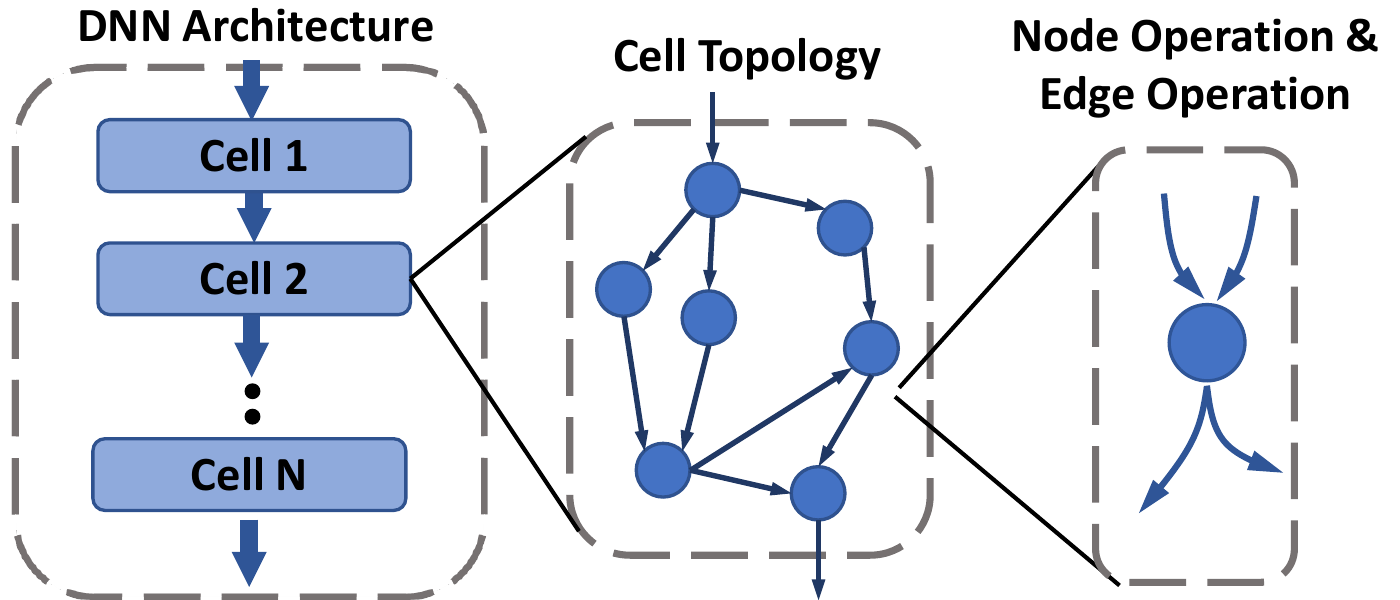}
\caption{Neural architecture structure and cell topology.
A DNN architecture is composed of several cell structures (Cell 1, Cell 2, ...). 
The \textit{cell topology} of a cell denotes the network connectivity patterns within the cell structure. 
For a \textit{cell topology}, operators are abstracted as nodes and tensor distributions are taken as edges, through which the tensors are communicated.}
\label{fig:hierarchy}
\end{figure}

Figure~\ref{fig:hierarchy} depicts the neural architecture hierarchical structure and its building block -- cell structures, the topology of which can be descried as node operations and the connectivity between nodes, i.e., edge operations.
When searching for more efficient neural architectures, the cell-based search approach is commonly employed~\cite{tan2019efficientnet,cai2018proxylessnas}.
These methods adopt the architecture motifs from hand-crafted models (e.g., MobileNetV2) as backbone structures. 
The found architectures have compact structure and compelling performance.
However, the cell-based approach heavily relies on existing cell structures and has constrained search space. 
It is not able to further discover the topology of an existing cell structure, which is
likely to induce performance degradation.

While the node-based approach does not depend on existing cell structures,
the topology-agnostic mechanism, i.e., pre-defined node graph topology, is utilized to optimize the corresponding operations~\cite{liu2018darts,pham2018efficient}.
It is impossible to fully explore the network connectivity patterns within cell structures.
The recently proposed randomly wired neural networks~\cite{xie2019exploring} utilizes a random graph priors to facilitating the interaction of tensors in DNN architectures.
However, its topology-agnostic nature makes the crafted models prone to structural redundancy and sub-optimal performance.

In this paper, we propose \as{}---a topology-aware NAS methodology.
By exploring the cell topology within cell structures, we aim to improve the performance and efficiency of found neural architecture 
and avoid the search space explosion due to the increased cell topology dimension. 
More specific, \as{} adopts a node-based search strategy by abstracting DNN operations as nodes in Directed Acyclic Graphs (DAGs) and the distribution of tensors as edges between nodes.
The search starts with a complete DAG and leverage its interconnected topology to fully utilize the flow of tensor between nodes. 
To reduce the risk of space explosion,
we introduce a topology knowledge accumulation mechanism and progressively optimize the cell structure through edge shrinking.
\as{} can explore the significantly larger and more flexible search space with less search time as the search space is intrinsically reduced with the shrinking of edges.

\as{} supports a wide range of applications, including Convolutional Neural Network (CNN)-based and Recurrent Neural Network (RNN)-based models.
We prototype \as{} and conduct a case study for crafting mobile-friendly neural architectures, where efficiency plays a critical role due to the highly constraint computing resources.
We evaluate \as{} for CNN and RNN architecture search on image classification and language tasks, respectively. 
ShrinkCNN that is crafted over the ImageNet-1K dataset~\cite{imagenet_cvpr09} has the similar accuracy performance as state-of-the-art (SOTA) techniques. 
Meanwhile, it has only $\mathrm{3.6M}$ parameters, 48\% or 25\% reduction compared to the hand-crafted MobileNetV2 ($\mathrm{6.9M}$)~\cite{sandler2018mobilenetv2} or MNasNet-A ($\mathrm{4.8M}$)~\cite{tan2019mnasnet}, respectively.
In terms of computational cost, ShrinkCNN cuts of 34\% MACs compared to MobileNetV2 while providing the similar accuracy. 
ShrinkRNN, a RNN model discovered using the Penn-Treebank dataset~\cite{marcus1994penn}, achieves competitive performance with SOTA models by taking only $1.5$ GPU hours search time, which is $6.7\times$ faster than ENAS \cite{pham2018efficient} and $16\times$ faster than the first-order DARTS \cite{liu2018darts}.

\section{Related Work}

\textbf{Neural Architecture Search (NAS)} promotes the design of SOTA and efficient neural architectures by exploring combinations of node operations, activation functions, etc.
Some existing works \cite{cai2018proxylessnas,wu2019fbnet,tan2019mnasnet} adopt the cell-based neural architecture search which reuses architecture motifs from hand-crafted architectures.
Despite such architecture motifs help reduce the search space, the fixed topology of 
existing cell structures severely restrict the utilization of information flow between nodes, which may lead to degraded performance of found architectures. 

Recent studies~\cite{pham2018efficient,liu2018darts} explore the
node-based approach, which relaxes the design space to a combination of node operations instead of constructing with predefined cells.
However, all these explorations take a topology-agnostic mechanism: 
the node graph topology is fixed before kicking off the search process and optimizing the corresponding operations.
As a result, structural and topological knowledge cannot be explored and accumulated during the search process, leading to sub-optimal performance and long search time.

\noindent\textbf{Wired Neural Architectures} is  proposed recently and attracted a lot of attention~\cite{xie2019exploring,wortsman2019discovering,cheng2019MSNet}. 
Random wiring~\cite{xie2019exploring} shows that random graphs generated by stochastic network generators can provide strong priors for network connectivity patterns.
The flow of tensors in such a random wiring allows more feature interaction and thus enhances the performance of found DNN architectures.
However, unconstrained random wiring may lead to an explosion of memory consumption due to the aggregation of tensors (e.g., addition and concatenation) in DNN operations.
More importantly, network generators can only learn the hyperparameters used to generate the random graphs (e.g., edge connection probability, degree sequence distribution of each node, etc.), rather than the actual knowledge of the graph topology.

\noindent\textbf{Network Morphism} attempts to morph the architecture of a neural network but keeps its functionality~\cite{wei2016network,gordon2018morphnet}. 
These methods mainly focus on morphing the depth, width, and kernel size of a network, 
or optimizing parallel towers in Inception-like DNN architectures~\cite{szegedy2016rethinking}. The knowledge of cell topology 
is hardly inherited.
Moreover, structure optimization based on network morphism can be time-consuming as it takes thousands of optimization steps to explore a competitive child network.

\section{Methodology}

\begin{figure*}[!t]
\centering
\includegraphics[width=1.8\columnwidth]{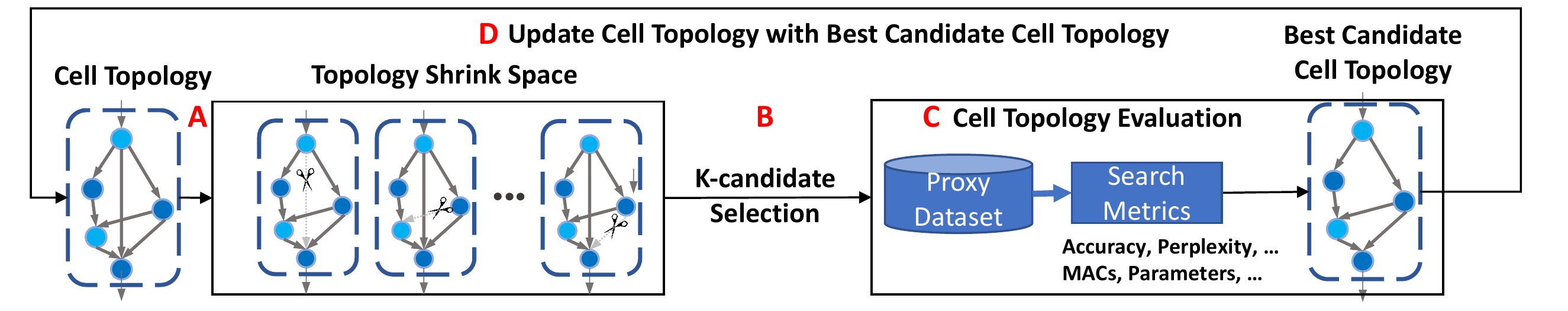}
\caption{An overview of \as{} workflow. 
\textbf{A:} Construct a \textit{topology shrink space} for a \textit{cell topology} by iteratively removing one edge. 
\textbf{B:} Randomly select $K$-candidate from the \textit{topology shrink space}. 
\textbf{C:} These selected candidates are evaluated by efficiency-aware search metric. \textbf{D:} The architecture that performs the best in evaluation is used to update the cell topology. 
The different node colors in a cell topology represent different operations.}
\label{fig:workflow}
\end{figure*}

\subsection{An Overview of \as{} Workflow}
Our proposed \as{} is a topology-aware NAS methodology for discovering efficient DNN architecture. 
More specifically, it explores the cell topology by progressively removing those redundant edges that do not contribute much to the model performance.
Such an approach indeed present a knowledge accumulation mechanism for the topology. 
As such, \as{} is able to search larger and more flexible space by paying reasonable search time.

Figure~\ref{fig:workflow} depicts the workflow of \as{} for a cell topology that is initially abstracted as a complete DAG.
We progressively optimize its structure through the following four phases iteratively.

$\bullet$ Phase A: For a cell topology, construct the \textit{topology shrink space} by removing one edge at a time and collecting all the derived candidates. 

$\bullet$ Phase B: Randomly select $K$ candidate cell structures from the \textit{topology shrink space} 
and formulate $K$ according to DNN architectures.

$\bullet$ Phase C: Collect the search metrics on the proxy dataset and use it to identify the best candidate cell topology from the $K$ randomly chosen candidates. 

$\bullet$ Phase D: Update the cell topology for the next iteration with the best one identified in Phase C. 
Because the new cell topology is derived by removing one edge from the previous iteration, the update process is called as \textit{edge shrinking}.

The \as{} workflow is an iterative process and eventually stops when no edge exists in the cell topology of interest. 
More details of the workflow is elaborated in Sections~\ref{sec:cell_topology}$\sim$\ref{subsection:cellshrink}.
The best candidate cell topologies obtained at the end of all the iterations will be collected and used for the post-shrink architecture construction, see Section~\ref{subsection:postshrink}.

\subsection{Cell Topology}
\label{sec:cell_topology}
The main optimization objective of the proposed \as{} is cell topology, which represents the network connectivity patterns of a cell structure. 
A cell topology is denoted as a DAG $G=(\mathcal{V}, \mathcal{E})$, 
where DNN operations are abstracted as {\it nodes} ($\mathcal{V}$) and the distribution of tensors are represented as {\it edges} ($\mathcal{E}$) connecting nodes.

A node operation denotes a DNN operation $o_v$ that processes the aggregation of tensors from other nodes and produces an output tensor $x_v$. 
$o_v$ is parameterized by its weight parameters $w_v$. 
It can be a convolutional layer when searching for CNN architectures, or a recurrent layer followed by a unique activation function, such as ReLU and tanh, when searching for RNN architectures.
Unlike random wiring~\cite{xie2019exploring} that assigns identical DNN operation for each node $v$, we assign each node $v \in \mathcal{V}$ a unique DNN operation $o_v$ to expand the search space.

\begin{figure}[t]
\centering
\subfigure[CNN node operation]{
\includegraphics[scale=0.32]{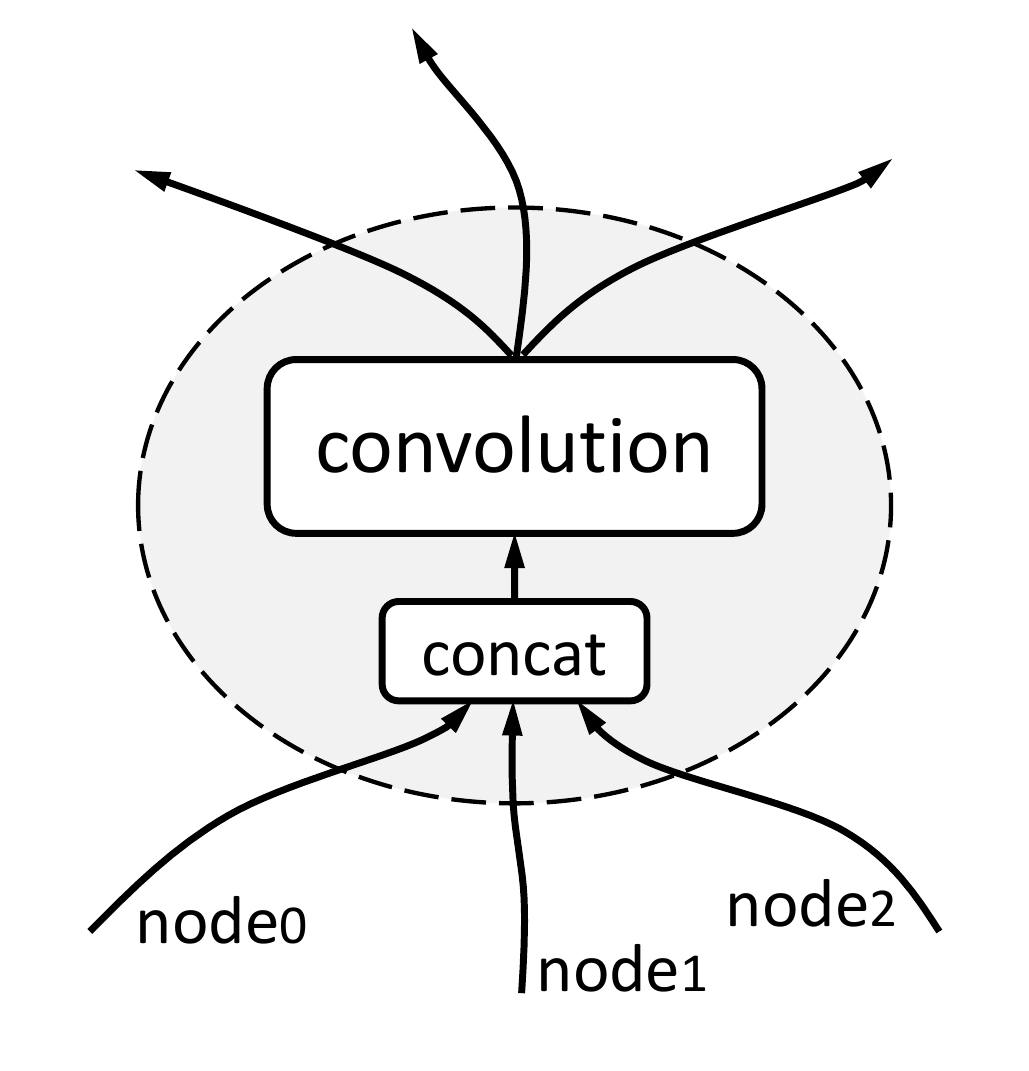}
\label{fig:cnn_operation}}
\subfigure[RNN node operation]{
\includegraphics[scale=0.32]{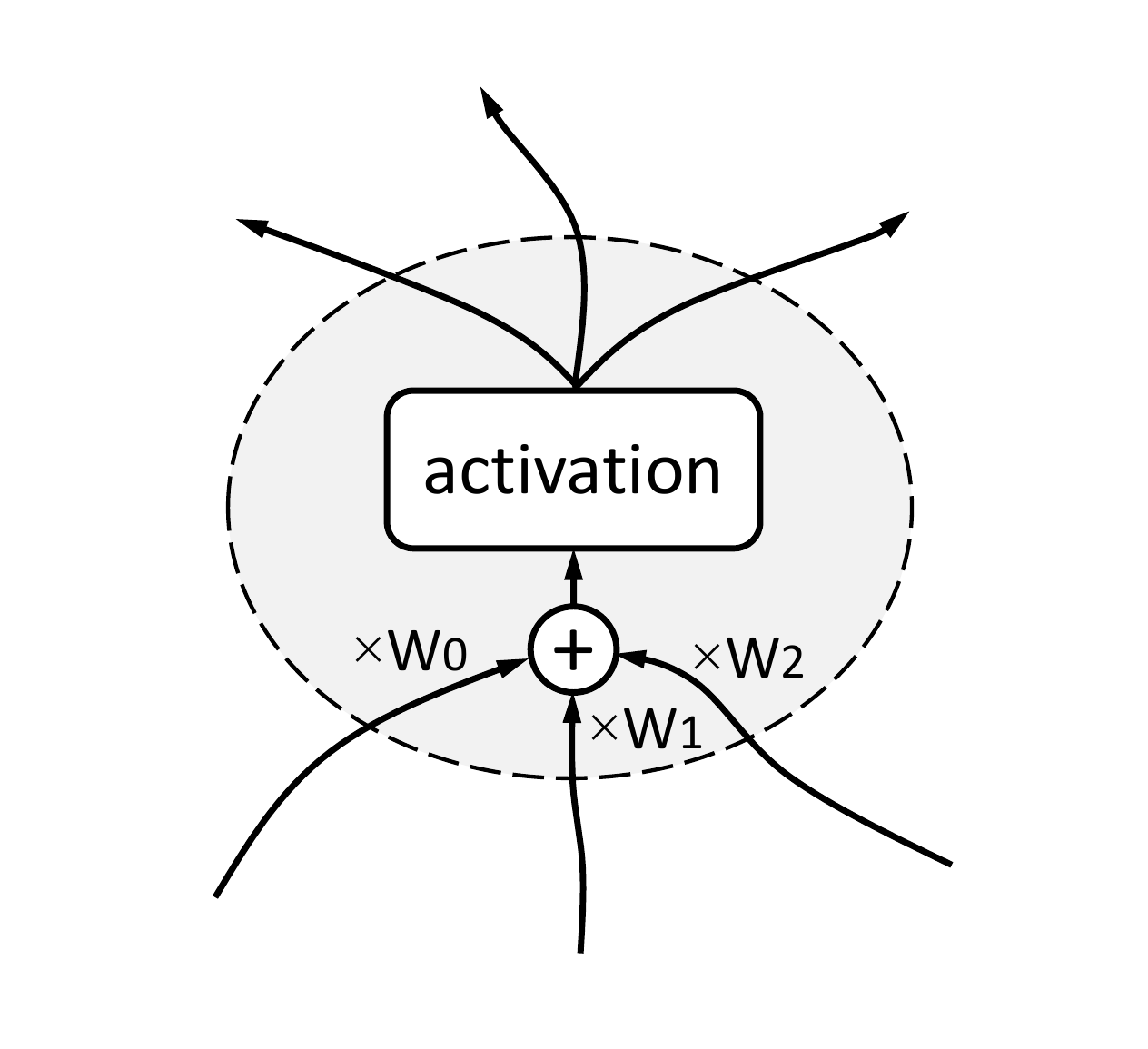}
\label{fig:rnn_operation}}
\caption{Node operations for CNN and RNN. Input nodes are first aggregated and processed by the DNN operation. The output will then be distributed to other connected nodes.}
\label{fig:node_operation}
\end{figure}

Figure~\ref{fig:cnn_operation} illustrates a CNN node operation. 
Existing works \cite{xie2019exploring} apply element-wise addition during the aggregation of tensors from different nodes.
To preserve the interaction of tensors, we use filter concatenation to aggregate tensors from different nodes and enable the flow of information between different layers.
The aggregated tensor is then processed by a convolution operation to produce an output tensor for the current node.
Given $k$ input tensors $x_{1}, x_{2}, ..., x_{k}$, the node $v$ computes the output tensors as:
\begin{equation}
x_v = o_v(Concat[x_{1}, x_{2}, ..., x_{k}]) ,
\end{equation}
where $Concat$ denotes the filter concatenation.

Figure \ref{fig:rnn_operation} gives an example of RNN node operation. 
Each of the input tensors is passed through a recurrent layer for a transformation.
The transformed tensors are then summed to form an aggregated tensor, which is passed through a non-linear activation function to produce the output tensor for this node.
Inspired by the memory mechanism in Recurrent Highway Networks (RHN)~\cite{zilly2017recurrent}, we use a highway bypass between adjacent nodes to get memory state.
Given $k$ input tensors $x_{1}, x_{2}, ..., x_{k}$, the node $v$ computes the output tensors as:
\begin{equation}
    c_i = \sigma(x_i\cdot w_i^1) ,
\end{equation}
\begin{equation}
    x_{v} = \sum_{i=1}^{k}c_i\otimes a(x_i\cdot w_i^2) + (1 - c_i)\otimes x_i ,
\end{equation}
where $\otimes$ denotes the element-wise multiplication, $w_i^1$ is highway gate parameters, and $w_i^2$ is transform matrices.
$\sigma$ denotes the $sigmoid$ activation function, and \textit{a} represents the assigned activation function for the current node.

A DAG that represents a cell topology can be directly mapped to an unique DNN building block.
During the mapping, a node that does not have any input connections are dropped. 
The output for the building block can be constructed from the leaf nodes with zero out-degree.
For CNN architectures, the leaf nodes are concatenated within the last dimension to produce an output feature map for the building block.
For RNN architectures, the leaf nodes are averaged to produce an output feature map.

\subsection{Topological Search Space Construction}

Random wiring \cite{xie2019exploring} uses identical DNN operation (e.g., a 3$\times$3 separable convolution) for all the nodes.
Our proposed \as{} removes this constraint and constructs a topological search space to enable the exploration of more flexible cell topology.

In CNN architecture search, each node can choose either $1\times1$ convolution or depth-wise separable $3\times3$ convolution as its operation. 
As batch normalization \cite{ioffe2015batch} speed up DNN training, every convolution operation adopts a Convolution-BatchNorm-ReLU triplet.
For RNN architectures, we extend the search space by randomly assigning unique non-linear function to each node following the recurrent layer. 
The non-linear functions include ReLU, sigmoid, tanh, and identity mapping.

\subsection{Edge Shrinking}
\label{subsection:cellshrink}

\as{} progressively optimizes the cell structure in a crafted topology shrink space.
In every iteration, \as{} compares the performance of candidate structures and accumulates topological knowledge.

\noindent\textbf{Topology shrink space} is used to describe the possible topological reduction during edge shrinking for improving the cell structure.
The topology shrink space $\pi$ is defined as a full set of all possible cell structures that can be derived by applying graph damage (i.e., removing one edge from the existing graph) to the current cell topology $g^{(t)}=(V^{(t)}, E^{(t)})$ at time $t$, such as
\begin{equation}
    \pi(g^{(t)}) = \{(V^{(t)},E^{(t)}-\{e\})|\forall e \in E^{(t)}\} .
\end{equation}

\noindent\textbf{Candidate cell structures.}
Considering the large topology shrink space induced by the node connection possibilities at time $t$, we adopt $K$-candidate selection strategy.
It randomly picks only $K$ candidate cell structures from the topology shrink space, and accumulates the topological knowledge from only the best candidate cell structure according to the search metrics.
This aggressive optimization immensely reduces the topology shrink space.
As our results in Section~\ref{sec:5.2} shall show, such an aggressive reduction in topology shrink space does not degrade much performance because the accumulated topological knowledge could compensate for the missing in the architecture evaluations. 
Despite of the reduction of topology shrink space, the intrinsic reduction on the overall search space is a leading factor to the lower search cost.
With the combination of the above two factors, the search cost of \as{} is significantly reduced.

\noindent\textbf{Shrink process.}
For a cell topology $g$, the shrink process targets to optimize it within the topology shrink space based on a resource-aware search metric:
\begin{equation}
S(g)={Perf(\mathcal{A}(g);\mathcal{D}) - \lambda \cdot\log{ Res(\mathcal{A}(g))}},
\label{eq:objective}
\end{equation}
where $\mathcal{A}(g)$ denotes the neural architecture built with $g$, and $\mathcal{D}$ is the proxy dataset used for evaluation.
$Perf()$ represents the best validation performance we can achieve on the proxy dataset. 
For CNN architectures for image classification tasks, \textit{accuracy} can be taken as the performance metric.
For RNN architectures targeting on language tasks, the performance metric can be \textit{perplexity}.
$Res()$ denotes the resource consumption of a model, such as the number of parameters or MACs of $A(g)$.
$\lambda$ is an adjustable parameter which penalizes the resource consumption to form a light-weight neural architecture. 

To search for efficiency-aware neural architectures, \as{} incorporates the resource-aware metric as a continuous penalty function into the search metric.
By evaluating the performance of candidate architectures in the topology shrink space and picking the best one according to the search metric, the current cell structure can better utilize the topological knowledge from similar structures and make improvement by adapting to the best candidate cell structure in the topology shrink space.
The pseudo code for the shrink process is given in Algorithm \ref{alg:autoshrink}.

\renewcommand{\algorithmicrequire}{ \textbf{Input:}}
\renewcommand{\algorithmicensure}{ \textbf{Output:}}
\begin{algorithm}[t]
\caption{\as{} 
}
\label{alg:autoshrink}
$\left. \textbf{Input:} \right.$ \newline
$\textbf{N}$: Number of nodes in the initial complete DAG. \newline
$\textbf{K}$: Number of candidates in topology shrink space to evaluate in each shrink step.\newline
\textbf{begin}
\begin{algorithmic}
\STATE Generate a complete graph $g^{(0)}$ with N nodes and randomly assign DNN operations.
\STATE $t \gets 0$
\WHILE {$E^{(t)}$  $\neq \{\emptyset\}$} 
    \STATE Phase A: Construct the topology shrink space $\pi(g^{(t)})$
    \STATE Phase B: Adopt K-candidate selection strategy to select K candidate cell structures $g^{(t)}_{1},g^{(t)}_{2}, ..., g^{(t)}_{K}$.
        \STATE Phase C: Construct the candidate DNNs using all candidate cell structures $g^{(t)}_{1},g^{(t)}_{2}, ..., g^{(t)}_{K}$.
        \STATE Train the candidate DNNs on proxy dataset and get the feedback search metrics  $S(g^{(t)}_{1}),S(g^{(t)}_{2}), ..., S(g^{(t)}_{K})$.. 
        \STATE Phase D: Use the best candidate cell structure to update cell topology.
        \STATE $g^{(t+1)}\gets \argmax_{g'\in \{g^{(t)}_{1},g^{(t)}_{2}, ..., g^{(t)}_{K}\} }S(g').$
\ENDWHILE
\RETURN \textbf{g}
\end{algorithmic}
\textbf{end}
\label{alg:autoshrink}
\end{algorithm}

\noindent\textbf{Topology knowledge accumulation.}
The progressive improvement process accumulates the topological knowledge in the cell structure as the edge shrinking steps forward.
The accumulation of topology knowledge is demonstrated in two aspects.
On the one hand, redundant edges are sequentially moved out of the cell topology to facilitate the exploration of efficient but representative cell topology.
On the other hand, due to the reduction of total number of edges in the cell topology, the search space is becoming smaller for the cell topology to make further improvement.
For example, the initial search space is estimated to contain $6.8\times10^{10}$ neural architectures for a given complete DAG with $28$ edges, $8$ nodes and $2$ possible choices for each node operation as the cell topology. 
After one shrink step, at most $27$ edges still exist in the cell topology and the search space is reduced by at least $4\times$, which contains at most $1.7\times10^{10}$ neural architectures.
Smaller search space enables a faster exploration of representative cell structure, which can be otherwise unexplored within limited time budget.

More formally, at time $t$, the progressive improvement of the current cell structure $g^{t}$ within the topology shrink space $\pi(g^{(t)})$ can be expressed as:
\begin{equation}
    g^{(t+1)} = \argmax_{g'\in \pi(g^{(t)})}S(g').
\end{equation}

\subsection{Crafting DNN Architecture}
\label{subsection:postshrink}

Based on the performance metric $S$ in Eq.~(\ref{eq:objective}), the cell structure with the best performance $g^{opt}$ will be taken as our representative cell structure of optimal performance:
\begin{equation}
    g^{opt} = \argmax_{\hat{g} \in \textbf{g}} S(\hat{g}).
\end{equation}
This representative cell structure is used as the building block to construct DNN architectures.

\begin{table}[b]
	\centering
	\scalebox{0.95}{
		\begin{tabular}{c|c|c}
			\hline
			Hierarchy & Output  & Regime \\
			& resolution & \\
			\hline
			Stem CONV & 32$\times$32 & CONV 3$\times$3 32 filters \\
			\hline
			MP + Stage 1     & 32$\times$32 & \as{} Cell, $T$, 16 filters \\
			\hline
			MP + Stage 2 & 16$\times$16 & \as{} Cell, $T$, 32 filters \\
			\hline
			
			MP + Stage 3 & 8$\times$8 & \as{} Cell, $T$, 64 filters\\
			\hline
			Classifier & 1$\times$1 & AP, FC, Softmax\\
			
			\hline
		\end{tabular}
	}
	\caption{The ShrinkCNN architecture configuration for the CIFAR-10 task. 
		CONV denotes the Convolution-BatchNorm-ReLU triplet. MP and AP denotes MaxPooling and Average Pooling respectively.
		On CIFAR-10 task, ShrinkCNN has 3 stages with $T$ \as{} cells stacked in each stage.}
	\label{tab:arch_cnn}
\end{table}

\noindent\textbf{Modularization for CNN.}
As CNNs are well-known to be a hierarchical design with different feature map size in different stages,
we divide a DNN architecture into multiple stages.
Instead of strided convolution, MaxPooling is used as the down-sampling module to connect two adjacent stages.
Table~\ref{tab:arch_cnn} gives an example of modularizing \as{} cells to construct CNN architectures for the CIFAR-10 task~\cite{krizhevsky2009learning}. 
In each stage, we stacked $T$ representative cells found by \as{} (i.e., $g^{opt}$).
The width of these cells remains the same within a stage, and doubles when passing the down-sample module.
Furthermore, a residual connection~\cite{he2016deep} is added from the input node to the output node within our optimal CNN cell structure, which is helpful to develop extra network connectivity patterns.
With the incorporation of residual connections, the performance grows with the depth of DNN architecture like ResNets~\cite{he2016deep}.

\noindent
\textbf{Modularization for RNN.}
To ensure the maximum flexibility in RNN architecture construction.
we do not assume any repetitive patterns while constructing optimal RNN architectures from optimal RNN cell structures.
The final RNN architecture for Penn Treebank consists of an embedding layer, an optimal RNN cell explored by \as{}, and a decoder to generate the final predictions.
Following DARTS~\cite{liu2018darts}, we use a fully connected layer to construct the decoder so that the hidden state can be used to predict.

\section{Experiments and Discussion}

We implement and evaluate \as{} for CNN and RNN architecture search respectively on image classification and language tasks. 
Considering there are $7\sim12$ nodes in SOTA node-based NAS~\cite{pham2018efficient,liu2018darts}, we set the initial cell structure to be a complete DAG with $N=8$ nodes for CNN search and $N=6$ for RNN search.  
Empirically, we set $\lambda$ to 0.1 to favor efficiency-aware architecture in both CNN and RNN search.

\subsection{\textit{ShrinkCNN} for Image Classification}
\noindent\textbf{Representative CNN cell structures from proxy dataset.}
We construct the proxy dataset for image classification tasks by randomly selecting 5,000 examples from the CIFAR-10 dataset~\cite{krizhevsky2009learning} with an equal distribution of classes.
To obtain representative cell structures, we first build candidate CNN architectures based on the 
candidate cell structures obtained from the \as{} process.
A candidate architecture follows the configuration in Table~\ref{tab:arch_cnn}: it has three stages; each stage contains one \as{} cell ($T=1$); and the numbers of convolutional filters in the three stages are 16, 32, and 64, respectively.

These candidate neural architectures are trained on the proxy dataset.
MAC is taken as the focused optimization resource and integrated into the search metrics. 
In each shrink step, we adopt a K-candidate selection strategy with $K$ set to 10 to deliver a fast architecture search without sacrificing the performance of cell topology.
Each training takes about 20 epochs to reach convergence.
We then evaluate the validation accuracy on the proxy dataset to get the feedback search metrics.
The largest candidate architecture derived from our proxy dataset has a computational cost of $\mathrm{20M}$ MACs and takes about 40 GPU seconds to reach convergence.

\begin{figure}[b]
\centering
\includegraphics[width=.95\columnwidth]{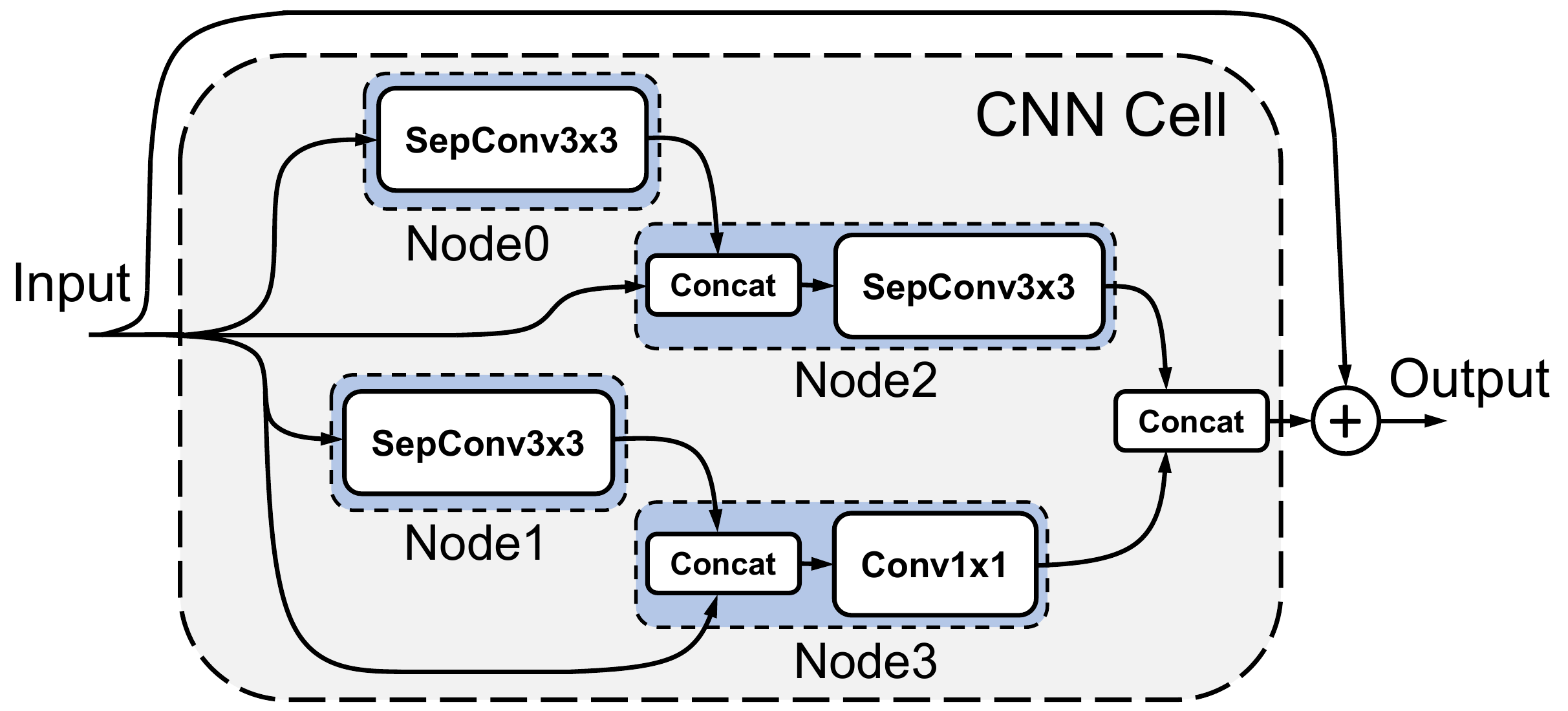}
\caption{The representative CNN cell structure.}
\label{fig:cnn_cell}
\end{figure}

Once the \as{} process is completed, the cell structure that provides the best result of search metric is selected as the representative cell structure.
As shown in Figure~\ref{fig:cnn_cell}, the representative CNN cell structure consists of 4 node operations and 3 filter concatenations.
The left part of the structure develops a Siamese pair of convolutions with wired concatenations to produce similar feature maps for the following different-sized convolutions.
The right part utilizes the previous input, and learns a combination of different-sized filters to capture the spatial information.

\noindent
\textbf{ShrinkCNN on ImageNet-1K.}
The representative CNN cell structure is adapted to ImageNet-1K dataset~\cite{imagenet_cvpr09} for crafting the optimal network architecture, namely, \textit{ShrinkCNN}.
In this work, we craft two architectures. 
ShrinkCNN-A is obtained by using ReLU activation, which can provide a fair comparison of \as{} and the ReLU-based CNN architectures~\cite{sandler2018mobilenetv2,liu2018darts,xie2019exploring}. 
ShrinkCNN-B is crafted by using swish~\cite{ramachandran2017searching} activation that helps increase the representation power of models. 
We compare ShrinkCNN-B with the swish-based CNN architectures~\cite{tan2019efficientnet,tan2019mnasnet}.

We adopt similar pre-processing pipeline as Inception-V3~\cite{szegedy2016rethinking} and use RMSprop optimizer \cite{hinton2012neural} with an initial learning rate 0.1 to optimize the CNN architectures.
The cosine learning decay suggested in SGDR~\cite{loshchilov2016sgdr} is employed to reduce the generalization error.

\begin{table}[t]
    \centering
    \begin{tabular}{|c|c|c|c|}
    \hline
         Architecture & Top-1 Error &  \# Param & MACs  \\
         &  &   (M)  & (G) \\
    \hline\hline
            ResNet-50 & \textbf{24.0\%} & 26 & 4.1 \\
            MorphNet & 24.8\% & 15.5 & -  \\
            MobileNetV1 & 29.4\% & 4.2 & 0.569 \\
            MobileNetV2 & 25.3\% & 6.9 & 0.585 \\
    \hline
            MnasNet-A & 24.4\% & 4.8 & \textbf{0.340} \\
            EfficientNet-B0 & 23.7\% & 5.3 & 0.391  \\
            DARTS & 26.7\% & 4.7 & 0.574 \\
            RandWire-WS & 25.3\% & 5.6 & 0.583 \\ 
    \hline
            ShrinkCNN-A & 26.1\% & \textbf{3.6} & 0.385 \\
            ShrinkCNN-B & 24.9\% & \textbf{3.6} & 0.385 \\
            \hline
    \end{tabular}
\caption{Performance comparison of various CNN architectures on ImageNet-1K dataset. 
	All the evaluations are based on 50,000 images of ImageNet-1K validation dataset. 
	The input resolution is set to $224\times224$.}
    \label{tab:imagenet}
\end{table}

\begin{table}[t]
\centering
\begin{tabular}{|c|c|c|c|}
\hline
Architecture & \underline{Perplexity} & \#Param & Search Cost\\
&  valid\quad test &  (M)  & (GPU hours)\\
\hline\hline
LSTM & 60.7\quad 58.8 & 24& - \\
LSTM-SC & 60.9\quad 58.3 & 24 & - \\
\hline
RHN & 67.9\quad 65.4 & 23 & - \\
\hline
ENAS  & \ \ -\quad\ \ \ \ 55.8& 23 & 10\\
\hline
DARTS  & 58.1\quad 55.7 & 23 & 24\\
\hline
ShrinkRNN & 58.5\quad 56.5  & 23 & 1.5\\
\hline
\end{tabular}
\caption{Performance comparison of various neural architectures on Penn Treebank dataset. We evaluate ShrinkRNN on both the validation and test dataset for fair comparison.}
\label{tab:ptb}
\end{table}

\noindent
\textbf{Performance evaluation.}
We compare ShrinkCNN-A and ShrinkCNN-B with SOTA hand-crafted and automatically searched models.
Table~\ref{tab:imagenet} summarizes their key performance metrics on the ImageNet-1K dataset.
In general, ShrinkCNN requires fewer parameters and MAC operations while providing the similar accuracy.
For example, compared to the hand-crafted MobileNetV2 model with ReLU activation, ShrinkCNN-A reduces 48\% model parameters and 34\% MACs. 
Compared to MNasNet-A which is crafted by conducting resource-aware neural architecture search, ShrinkCNN-B can further cut off 25\% parameters with negligible impact on the top-1 error rate.
Our ShrinkCNN-B requires slightly more MACs mainly because MnasNet-A applies architecture motifs (e.g., squeeze-and-excitation layers) to further increase the efficiency of MACs. Such architecture motifs can also be combined with ShrinkCNN-B to further boost the performance. Here we only show the performance results of \as{} without prior knowledge to demonstrate its effectiveness.
Compared to EfficientNet \cite{tan2019efficientnet} obtained by scaling up MobileNetV2 blocks using automated search,
ShrinkCNN-B can save 32\% parameters with the similar top-1 error rate and MACs.

\subsection{\textit{ShrinkRNN} for Language Tasks}
\noindent\textbf{Representative RNN cell structures from proxy dataset.}
In this work, the proxy dataset for language tasks is constructed by randomly selecting 4,000 sentences from the Penn Treebank dataset.
We adopt the similar approach for CNN architectures in the search of representative RNN cell structure. 
Specifically, for language tasks, we set the dimension of embedding and hidden units of candidate cell structures to $200$~\cite{liu2018darts,pham2018efficient} and incorporate the number of parameters as efficiency-aware resource consumption into the search metric.
In each shrink step, we adopt a K-candidate selection strategy with $K$ set to 5 to balance the search speed and the performance of cell topology.
The training of candidate neural architecture on the proxy dataset takes about 10 epochs to converge.
The largest RNN architecture derived from our candidate cell structures has $1.2$ Million parameters and takes about 80 GPU seconds to reach convergence.

\begin{figure}[t]
\centering
\includegraphics[width=.95\columnwidth]{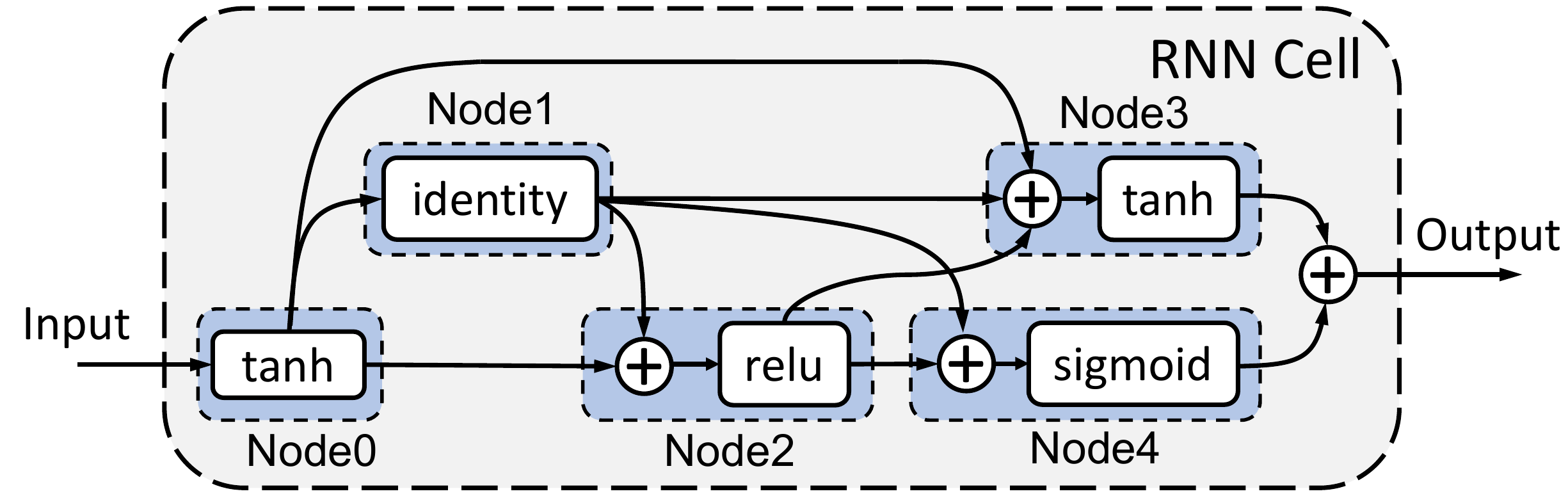}
\caption{The representative RNN cell structure.}
\label{fig:rnn_cell}
\end{figure}

Figure~\ref{fig:rnn_cell} depicts the found representative RNN cell structure.
It has 5 node operations and 4 node additions.
The left node combines the hidden state from the previous step and the input word embedding. 
Each node on the right side learns a combination of others' outputs to capture high-level information.
This RNN cell integrates various activation functions, which improves its representational power.

\noindent\textbf{ShrinkRNN on Penn Treebank.}
We adapt our representative RNN cell structure to the full Penn-Treebank dataset~\cite{marcus1994penn} for crafting \textit{ShrinkRNN}. 
For comparison purpose, we rescale ShrinkRNN to a fixed setting of 23 Million parameters.
NT-ASGD algorithm~\cite{merity2018regularizing} is used to train the ShrinkRNN architecture and the initial learning rate is set to 20.
Additional regularization techniques include an $\ell_2$ regularization weighted by $8\times 10^{-7} $; variational dropout~\cite{NIPS2016_6241} of 0.2 to word embeddings, 0.75 to cell input, 0.25 to hidden nodes and 0.75 to output layer.

\noindent
\textbf{Performance evaluation.}
We compare ShrinkRNN with SOTA models including LSTM~\cite{merity2018regularizing}, LSTM with skip connections (LSTM-SC)~\cite{melis_lstm}, recurrent highway network (RHN)~\cite{zilly2017recurrent}, ENAS~\cite{pham2018efficient}, and DARTS~\cite{liu2018darts}.
Table~\ref{tab:ptb} summarizes the validation and test results on the Penn-Treebank dataset. 
ShrinkRNN provides a comparable performance while its crafting time is significantly shorter (over $6.7\times$ shorter) than the existing node-based NAS methods.

\subsection{Ablation Studies}
\label{sec:5.2}
\noindent\textbf{Efficiency of CNN cell topology.}
We analyze the progressive optimization procedure towards our representative CNN cell structure to evaluate the efficiency of our cell topology.
Starting with a complete DAG with $N=8$ nodes and 28 edges, \as{} progressively optimizes the CNN cell topology to only 6 edges.
The computational cost of our representative cell structure reduces from $\mathrm{6.12M}$ MACs to $\mathrm{2.60M}$ MACs.
To verify the effectiveness of our knowledge accumulation mechanism, 
we compare the performance of neural architectures constructed from complete-DAG-based cell topology and strong topology priors with our representative cell topology.
The topology priors we use are Watts-Strogatz (WS) graph, Erd\"{o}s-R\'{e}nyi (ER) graph, and Barab\'{a}si-Albert (BA) graph, which are popular in graph theory. 

We craft CNN architectures with the above topology respectively, train them on the full CIFAR-10 dataset with the same hyperparameter setting, and compare their test results with ShrinkCNN in Table \ref{tab:efficent_topo}.
\begin{table}[b]
\centering
\begin{tabular}{|c|c|c|c|c}
	\hline
	Topology & Nodes & Accuracy & MAC \\
	&       &  (\%)    &   (M) \\
	\hline
	Complete-DAG & 8 & 91.56 & 105.61\\ 
	WS & 15 & 90.5$\pm$0.41 & 16.04$\pm$2.71 \\
    ER & 15 & 92.1$\pm$0.38 & 34.40$\pm$16.43 \\
    BA & 15 & 90.0$\pm$0.34& 32.61$\pm$5.96\\
    ShrinkCNN & 8 & 93.22 & 38.20 \\ 
	\hline
\end{tabular}
\caption{Comparison with ShrinkCNN topology with prior graph topology. We use $N=8$ nodes when crafting CNN architectures with complete-DAG-based topology, and $N=15$ nodes when crafting CNN architectures with random graph priors. Experiments on random graph priors are conducted 10 times to reduce randomness as much as possible.}
\label{tab:efficent_topo}
\end{table}
Table \ref{tab:efficent_topo} shows that ShrinkCNN is able to achieve up to 1.66\% higher accuracy than the complete-DAG-based topology, and up to 3\% higher than the topology from random graphs, thanks to the topological knowledge accumulation in the \as{} process.

\noindent\textbf{Efficiency of RNN cell topology.}
We also analyze the progressive optimization procedure in RNN cell structure. 
Starting with complete DAG with 6 nodes and 15 edges, \as{} progressively optimize the cell topology to only 8 edges, while the number of parameters within the cell topology is reduced from $\mathrm{3.37M}$ to $\mathrm{2.8M}$.
We craft RNN architectures using complete-DAG-based topology and our representative cell structures.
Then, we train the RNN architectures on full Penn Treebank dataset with the same hyperparameters.
Table \ref{tab:rnn_dag} indicates that our representative RNN cell structure can achieve a significantly lower (6.6) perplexity with 30\% fewer parameters.

\begin{table}[t]
\centering
\begin{tabular}{|c|c|c|c|}
	\hline
	Cell Topology & Edges & Perplexity  & \# Param (M) \\
	\hline
	Complete-DAG & 15 & 63.1  & 33 \\ 
	ShrinkRNN & 8 & 56.5 &  23 \\ 
	\hline
\end{tabular}
\caption{Demonstration of efficient RNN cell structures found by \as{} algorithm on full Penn Treebank dataset.
We used $N=6$ nodes when crafting RNN architectures with complete-DAG-based topology.}
\label{tab:rnn_dag}
\end{table}

\begin{figure}[h]
\centering
\includegraphics[width=0.95\columnwidth]{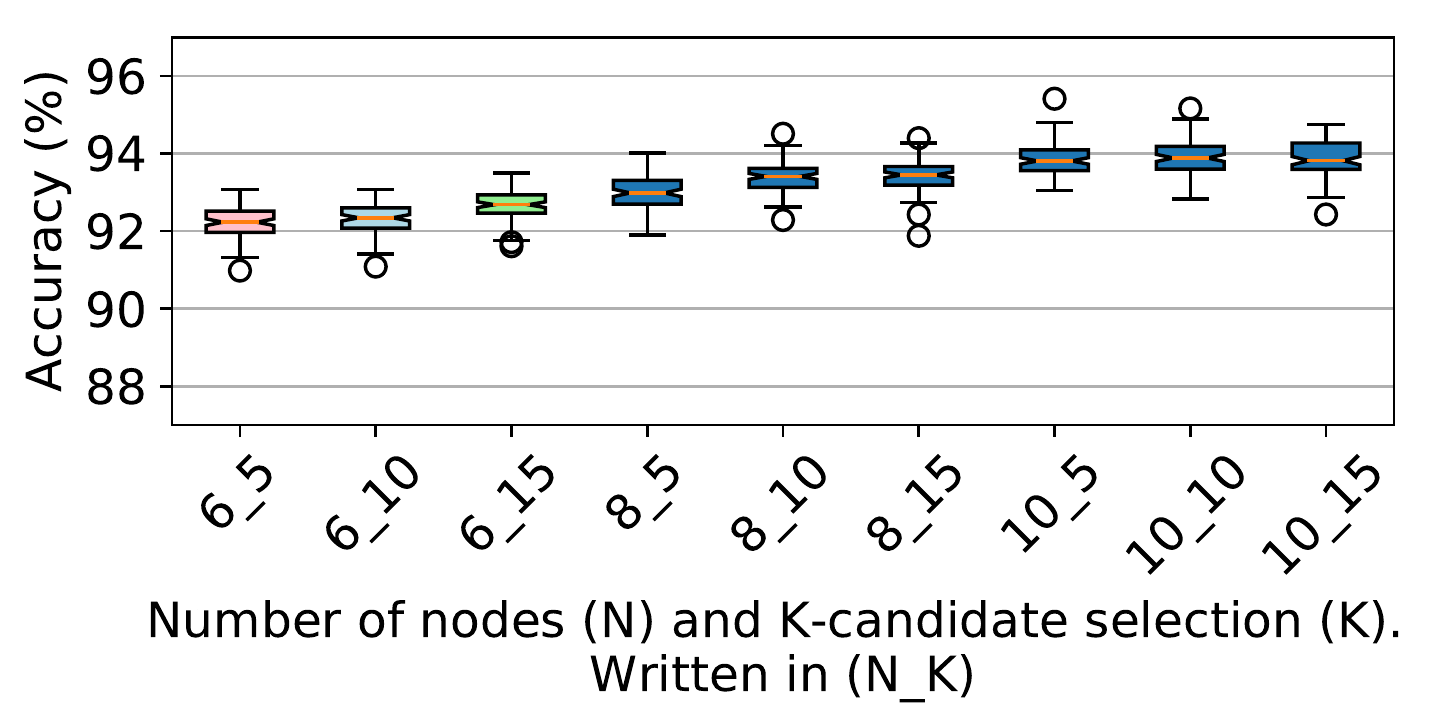}
\caption{The impact of $K$ on the performance of our representative CNN structures with different number of nodes at the beginning of \as{} process. Marginal performance gain is spotted when increasing $K$ to conduct more delicate \as{} process with larger search cost.}
\label{figure:topology_shrink_space}
\end{figure}

\noindent\textbf{K-candidate selection in CNN.}
We further investigate the impact of $K$ value in the $K$-candidate selection strategy and analyze its impact on both search cost and performance.
Specifically, we configure $K$ to be 5, 10, 15 while running \as{} on complete DAGs with 6, 8, 10 nodes.
From figure \ref{figure:topology_shrink_space}, we observe that different combinations of the number of nodes and $K$ candidates may result in slightly difference in final accuracy,
although increasing $K$ can strengthen the candidate selection process.
However, the search cost dramatically increases with $K$. 
For example, when $N=8$, changing $K$ from 10 to 15 increases the searching time from 1.5 hours to 1.875 hours, while the validation accuracy has subtle difference.

\section{Conclusion}
In this work, we propose a topology-aware NAS method, \as{}, for discovering efficient and representative cell topology to formulate DNN architectures.
To explore significantly larger and more flexible space, \as{} accumulates topology knowledge by combining intrinsic search space reduction and K-candidate selection strategy to significantly reduce the search cost. 
The above contributions make \as{} consistently outperform both hand-crafted models and automatically searched models on both image classification and language tasks without any prior knowledge.
Using representative cells found by \as{}, the crafted ShrinkCNN achieves comparable accuracy on ImageNet-1K dataset with up to 48\% parameter reduction and up to 34\% MACs reduction compared to SOTA models.
The crafted ShrinkRNN has comparable performance results as SOTA models while can be crafted 6.7$\times$ faster.

\paragraph{Acknowledgements.} This project is in part supported by the following grants: NSF-1937435, NSF CCF-1725456, DOE SC0017030, Qualcomm Gift, NSF CCF-1756013, IIS-1838024 (using resources provided by Amazon Web Services as part of the NSF BIGDATA program).
We also thank USTC for the support of computing.

\bibliography{references}
\bibliographystyle{aaai}

\end{document}